%% file: main.tex
\documentclass{article}

\usepackage[english]{babel}

\usepackage{tabularx}
\usepackage{supertabular}
\usepackage{multirow}
\usepackage{booktabs}
\usepackage{subcaption}

\usepackage[letterpaper,top=2cm,bottom=2cm,left=3cm,right=3cm,marginparwidth=1.75cm]{geometry}

\usepackage{amsmath}
\usepackage{graphicx}
\usepackage[colorlinks=true, allcolors=blue]{hyperref}

\title{Improving OCR for \\ Historical Texts of Multiple Languages}
\author{Hylke Westerdijk ~\;~ Ben Blankenborg ~\;~ Khondoker Ittehadul Islam\\[5pt] University of Groningen \\[5pt] {\tt \{h.p.westerdijk,  b.b.g.blankenborg, k.i.islam\}@student.rug.nl}\\}
\date{}

\begin{document}
\maketitle

\begin{abstract}
This paper presents our methodology and findings from three tasks across Optical Character Recognition (OCR) and Document Layout Analysis using advanced deep learning techniques. First, for the historical Hebrew fragments of the Dead Sea Scrolls, we enhanced our dataset through extensive data augmentation and employed the Kraken and TrOCR models to improve character recognition. In our analysis of 16th to 18th-century meeting resolutions task, we utilized a Convolutional Recurrent Neural Network (CRNN) that integrated DeepLabV3+ for semantic segmentation with a Bidirectional LSTM, incorporating confidence-based pseudolabeling to refine our model. Finally, for modern English handwriting recognition task, we applied a CRNN with a ResNet34 encoder, trained using the Connectionist Temporal Classification (CTC) loss function to effectively capture sequential dependencies. This report offers valuable insights and suggests potential directions for future research.
\end{abstract}

\section{Introduction}

The task of recognizing characters in handwritten text poses significant challenges. Though the fundamental shapes of letters remain consistent, each individual's unique writing style introduces variability. Additionally, the condition of the writing surface may deteriorate over time, and the absence of contextual clues can lead to ambiguity in interpretation. The process of analyzing extensive volumes of handwritten documents and converting them into a digital format demands considerable time and effort from human transcribers.

Automating this transcription process offers clear advantages. The creation of hand-engineered features for letter recognition proves to be a complex endeavor since the shapes of letters are influenced by numerous factors, including the writer's style, the type of pen used, the surface material, and the direction of writing. This complexity leads us naturally to consider Deep Learning, which employs hierarchical models capable of autonomously learning features and tasks from data. Furthermore, Deep Learning models typically demonstrate superior performance compared to those based on handcrafted features \cite{DL}. However, a notable limitation of Deep Learning is its dependence on a substantial volume of data for effective feature learning; insufficient data can result in overfitting, where the model excels only on the training set \cite{DL}. This issue becomes particularly pronounced in Optical Character Recognition (OCR) tasks involving historical handwritten manuscripts. Since manuscripts from centuries past are often scarce and varied in script, the challenge of obtaining adequate training data becomes even more significant.

In this report, we address three specific challenges related to Optical Character Recognition (OCR) through the application of Deep Learning methodologies. We begin by outlining the task description in this section. In Section \ref{sec:lit-review}, we review existing methods that have proven effective in similar tasks. Section \ref{sec:dataset} provides a detailed overview of the datasets used in our study. In Section \ref{sec:method}, we describe the methodologies implemented, and Section \ref{sec:exp-setup} outlines the experimental setup employed throughout the research. Section \ref{sec:results} presents our findings, while Sections \ref{sec:discuss} and \ref{sec:conclussion} offer our discussions and conclusion, respectively.

\subsection{Task 1}
The objective of this task is to conduct Optical Character Recognition (OCR) on a Hebrew subsection of the Dead Sea Scrolls dataset, which is comprised of fragments of handwriting dating from the third century BC to the first century AD \cite{Dead_sea_scrolls}. In order to effectively address this task, we initiated a data augmentation process to expand our training dataset. Subsequently, we employed Kraken \cite{kraken} and TrOCR \cite{TrOCR} models tailored for performing OCR on historical manuscripts, to train our model.

\subsection{Task 2}
In Task 2, we shift our focus from Optical Character Recognition to Document Layout Analysis. The primary goal of Document Layout Analysis is to uncover the underlying structure of the document by identifying various text types, such as paragraphs and headings, and classifying them accurately \cite{DLA}. For this analysis, we utilized the Staten of Overijssel dataset, which consists of meeting resolutions in Dutch, encompassing nine distinct text types. We were provided with 90 labeled images of this dataset, which serve as the foundation for our study.

Our methodology adopted a semantic segmentation approach, where each pixel in the image is assigned a classification, as opposed to an object detection approach that would delineate the boundaries of text regions prior to classification. First, we preprocessed the image to achieve attain better resolution quality. Later, we trained a DeepLabV3+ model \cite{DeepLabV3+} utilizing the limited available labeled images. Consequently, we applied our trained model for pseudolabeling the remaining images in the dataset. The images that received high-confidence pseudolabels were then used to enhance our model's training. Upon evaluating our model on the test set, we achieved a loss of 0.7030 and a mean intersection over union of 0.6506, indicating a solid performance in the classification task.

\subsection{Task 3}

Task 3 focuses on the accurate recognition of handwriting within the IAM dataset \cite{iam}, which comprises lines of English handwriting. For this task, we employed a Convolutional Recurrent Neural Network (CRNN) \cite{CRNN} integrated with a ResNet34 \cite{koonce2021resnet} backbone, which facilitates robust feature extraction from visual data. Following the feature extraction, we utilized a multi-layer BiLSTM \cite{bi_lstm} architecture to effectively capture sequential dependencies present in the handwriting. The training of this model was conducted using the Connectionist Temporal Classification (CTC) loss function to optimize performance.

\section{Literature review}
\label{sec:lit-review}

\subsection{Task 1}
% Laventish distance: when it is used and what it is in brief

The domain of OCR offers several methodologies, with a particular emphasis on Deep Learning-based approaches in this section. We begin by examining transformer-based techniques, such as Transformer-based Optical Character Recognition (TrOCR) \cite{TrOCR}. The TrOCR stands out as a leading method for handwriting recognition, demonstrating superior performance compared to existing state-of-the-art approaches for both printed and handwritten text. This model leverages a Transformer \cite{vaswani2017attention} architecture equipped with self-attention mechanisms, trained specifically using the Connectionist Temporal Classification (CTC) \cite{graves2012connectionist} loss function. Importantly, TrOCR does not rely on a Convolutional Neural Network (CNN) backbone, which can introduce biases. The architecture comprises an encoder that utilizes pre-trained Vision Transformer models to generate representations of image patches, while the decoder employs pre-trained large language models to decode the textual content derived from these image representations. The adoption of pre-trained models is crucial, particularly when the available data is limited, as transformer architectures typically require a substantial volume—often around one hundred thousand data points—to fully learn without prior biases \cite{transformerdata}. In this context, TrOCR employs DeIT \cite{touvron2022deit} and Beit \cite{bao2021beit} as pre-trained Vision Transformers, alongside RoBERTa \cite{roberta} and MiniLM \cite{minilm} as its language models, both of which have been trained on English data. This reliance on English-language models may present a challenge for recognition tasks involving different languages and scripts.

Another notable model, Handwritten Text Recognition with Vision Transformer (HTR-VT) \cite{HTR-VT}, addresses the data-intensive nature of vision transformers by incorporating a Convolutional Neural Network for feature extraction rather than relying solely on standard patch embedding. This model utilizes only the encoder component of the transformer architecture. Additionally, it introduces the span-mask technique, which serves as an effective regularization method by masking connected features within the feature map. Remarkably, HTR-VT has shown competitive results, achieving performance on par with or sometimes exceeding that of CNN-based models, even on smaller datasets of approximately 20,000 data points.

Lastly, we investigate Kraken \cite{kraken}, which employs a distinct approach by integrating a Convolutional Neural Network with a Bidirectional LSTM, trained with the CTC loss function, while omitting the transformer architecture altogether. This model is specifically designed for character recognition in historical documents, which are often more degraded than contemporary texts. Notably, Kraken exhibits impressive performance across various scripts, achieving a mean character accuracy exceeding 95\% on diverse scripts, ranging from Cyrillic to Latin. Particularly relevant for our application to the Dead Sea Scrolls Dataset, it attained a mean character accuracy of 96.9% on Hebrew script, highlighting its potential effectiveness for our objectives.

\subsection{Task 2}

Document Layout Analysis models can generally be categorized into two primary approaches: semantic segmentation methods and object localization methods. In this subsection, we will explore examples representative of both categories. 

A prevalent choice for object localization is the Mask R-CNN model. This method builds upon the Faster R-CNN framework \cite{faster_r_cnn} by incorporating an additional branch that generates segmentation masks for the proposed Regions of Interest. To ensure congruence between pixel locations in the segmentation masks and the Regions of Interest, Mask R-CNN employs RoIAlign instead of the conventional RoIPool. This segmentation branch operates concurrently with the bounding box regression and classification processes inherent to Faster R-CNN, effectively decoupling segmentation from classification. The modifications made to Faster R-CNN entail feeding feature maps produced by convolutional layers into a Region Proposal Network, which then generates Region Proposals that are subsequently combined with the feature maps to yield a consolidated feature map containing overlaid Region Proposals. When applied to newspaper layouts for document analysis, Mask R-CNN achieves a mean average precision of 81.6, accompanied by a mask loss of 0.13 \cite{mask_rcnn_instance_seg}.

In contrast to object localization, semantic segmentation classifies each pixel in an image according to its corresponding class, rather than merely generating a bounding box. An enhancement over the prior DeepLabV3 \cite{DeepLabV3} model, DeepLabV3 plus \cite{DeepLabV3+} introduces a decoder aimed at refining segmentation outcomes, resulting in an end-to-end convolutional neural network (CNN) comprised of both an encoder and decoder. This architecture has produced state-of-the-art results across several benchmark datasets. A frequently utilized backbone for the DeepLabV3+ encoder is ResNet50 \cite{resnet50}, which is a robust CNN architecture consisting of 50 layers and is pre-trained on ImageNet.

\subsection{Task 3}

OCR for a modern English handwriting dataset containing over 10,000 data points can be approached in various ways. TrOCR \cite{TrOCR} has demonstrated impressive performance on the IAM dataset, leveraging a transformer-based architecture. Another promising approach involves a vertical attention module tailored for end-to-end handwriting recognition of paragraphs or lines \cite{VerticalAttentionOCR}. This vertical attention module integrates hybrid attention mechanisms—both content-based and location-based—addressing the challenges of training a transformer from scratch with fewer than 100,000 data points. By applying data augmentations exclusively during training, this method generates 'new' data points in each epoch through slight alterations to the original data, such as perspective transformations or elastic distortions. This innovative module has resulted in state-of-the-art performance, achieving a character error rate of 1.91\% on the RIMES \cite{Grosicki2024-za} dataset. Nevertheless, it is noteworthy that the model requires a lengthy convergence period, typically around a thousand epochs.

Beyond standard text recognition, scene text recognition focuses on interpreting text captured in outdoor environments, where the text appears against varied backgrounds and amidst other objects. IterVM \cite{itervm} exemplifies a model that achieves State Of The Art outcomes by combining an iterative visual model with an iterative language model for optimal performance. However, the efficacy of such a model in conventional OCR tasks remains uncertain, particularly due to the added complexity introduced by variations in handwriting styles.

In addition to transformer methodologies, hybrid approaches utilizing Recurrent Neural Networks (RNNs) and Convolutional Neural Networks (CNNs) can be effective. For instance, a Bidirectional LSTM possesses the advantage of capturing more sequential information compared to its unidirectional counterpart \cite{bi_lstm}, making it particularly adept at understanding dependencies within a line of text. However, a Bidirectional LSTM cannot directly process image data; hence, it is paired with a Convolutional network to extract relevant features that are subsequently fed into the RNN. In \cite{CRNN}, one such Convolutional Recurrent Neural Network (CRNN) achieved an overall accuracy of 92\% in text recognition tasks. Notably, CRNNs leverage existing frameworks, enabling faster convergence and reduced data requirements compared to transformer-based models.

\section{Datasets}
\label{sec:dataset}
% In this section, try talking about what we were provided by the task and their characteristic.
%% For example, on task 1, we were provided with 10(?) different variation of a single hebrew character etc.,
%% Grey images

% On the methadology, we are writing what we have done with those sets.
\subsection{Task 1}

\begin{figure}
    \centering
    \includegraphics[width=0.5\linewidth]{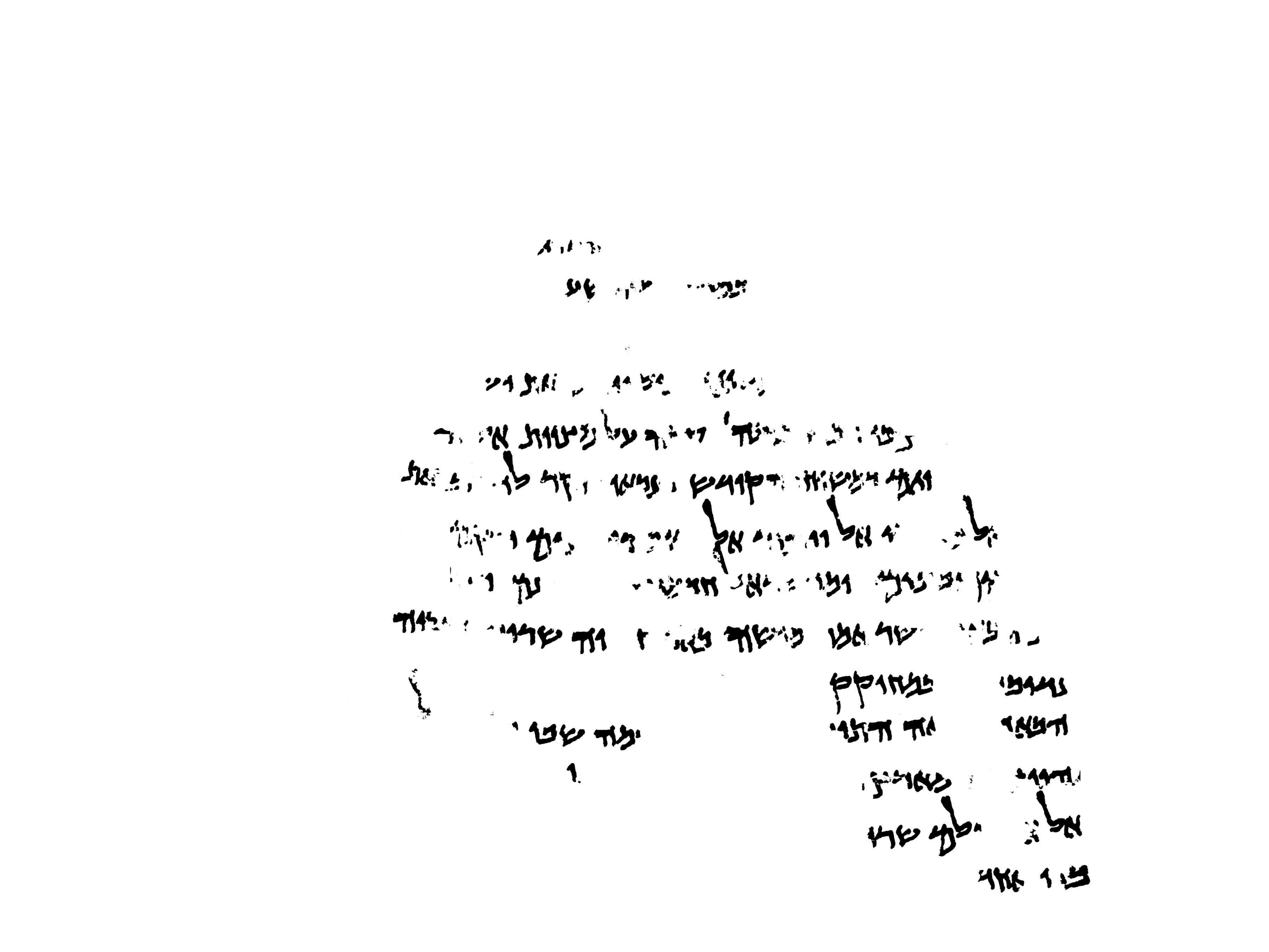}
    \caption{An binarized image from the dataset containing the text taken from the Dead Sea Scrolls.}
    \label{fig:bin_dss}
\end{figure}

For Task 1, we were provided with a selection of Hebrew pages from the Dead Sea Scrolls. It is important to note that Hebrew is read from right to left, which contrasts with the left-to-right structure that many memory-based models are typically trained on. The scrolls, dating back to the third century BC, exhibit significant material decay, resulting in damaged text that complicates character recognition tasks. We were provided with a total of twenty images, each offered in three distinct formats: RGB, grayscale, and binarized. For illustration, an example of a binarized image can be found in figure \ref{fig:bin_dss}. Additionally, we received various example images corresponding to each of the 27 characters derived from the 22 letters of the Hebrew alphabet along with their variations. 

\subsection{Task 2}
% Try saying about the resolution challenge which this dataset contains something we mitigated through double-to-single page separation

\begin{figure}
    \centering
    \includegraphics[width=0.5\linewidth]{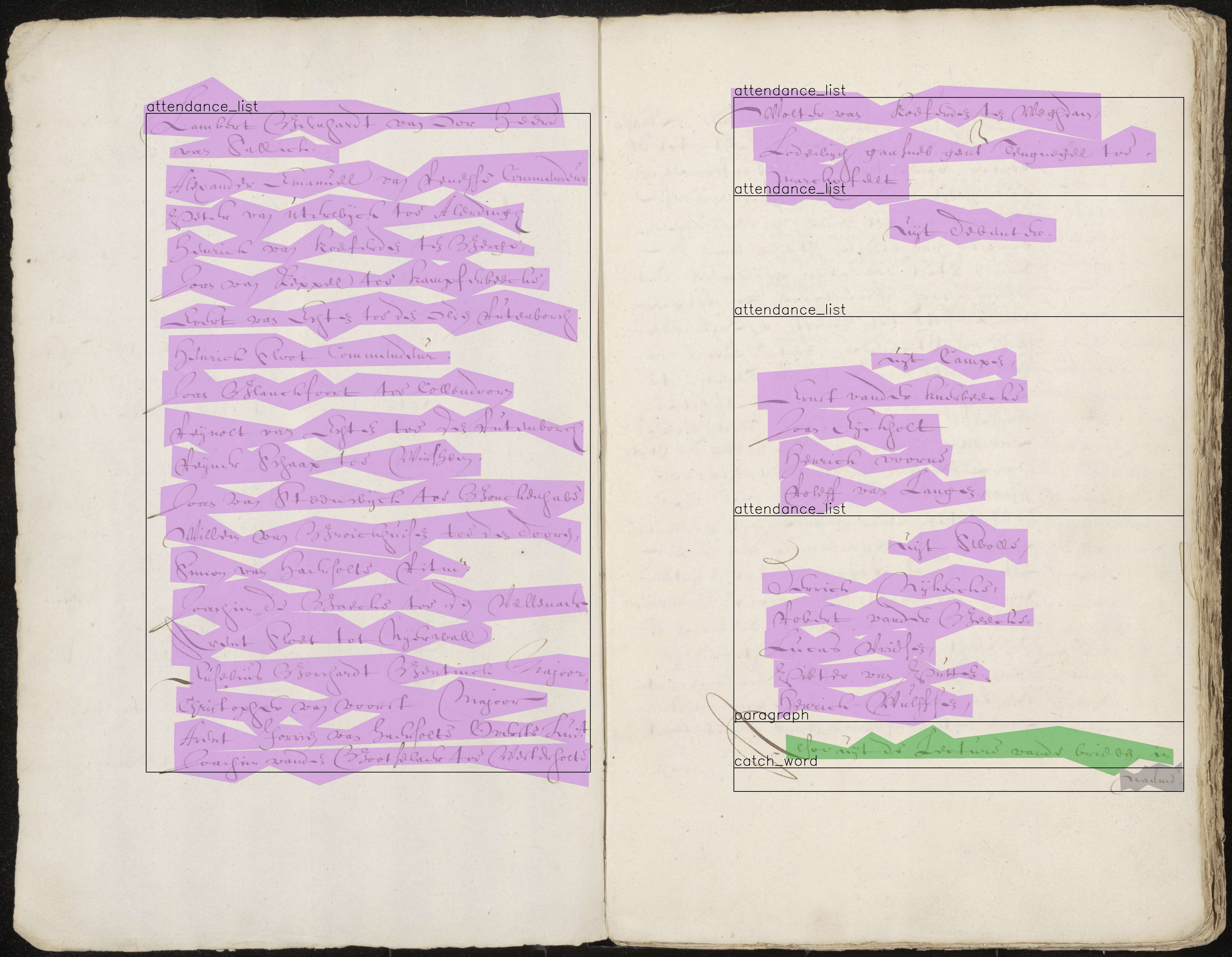}
    \caption{A resolutions page with both the segmentation mask and bounding box annotation.}
    \label{fig:staten_van}
\end{figure}

For Task 2, we were provided with a set of 90 images and corresponding labels derived from a collection of historical texts titled “Staten van Overijssel, Ridderschap en Steden en van de op hen volgende colleges.” These documents encompass resolutions from governmental meetings that took place between the years 1578 and 1795. Each image is accompanied by two distinct labels: a JSON file that details the bounding box coordinates and identifies the text regions, and a semantic mask file for each image, which assigns different pixel values to various classes. The images vary in size; some depict a single page, while others illustrate double pages. Within the semantic mask, different text types are categorized, each associated with a unique identifier: background is labeled as 0, headings receive a value of 1, paragraphs are assigned 2, requests (summaries) hold the value of 3, decisions are marked with 4, marginalia (notes) is labeled as 5, attendance lists are categorized as 6, catch words are noted as 7, and dates are represented with a value of 8. Additionally, there exists an 'undefined' category, marked as 11, which appears only once and lacks a clear interpretation. An example illustrating a page along with both the segmentation mask and bounding box overlay is presented in Figure \ref{fig:staten_van}. Notably, the distribution of these labels is imbalanced, with certain classes, such as paragraphs, occurring with significantly greater frequency than others, like dates.

\subsection{Task 3}

\begin{figure}[htb]
    \centering
    \begin{subfigure}[b]{\textwidth}
        \centering
        \includegraphics[width=0.99\linewidth]{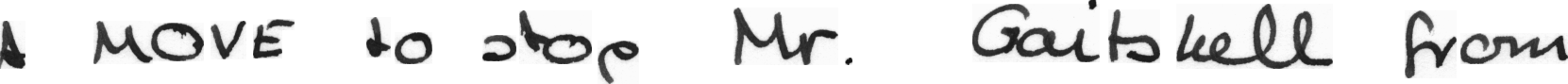}%
    \end{subfigure}
    \vskip\baselineskip
    \begin{subfigure}[b]{\textwidth}
        \centering
        \includegraphics[width=0.99\linewidth]{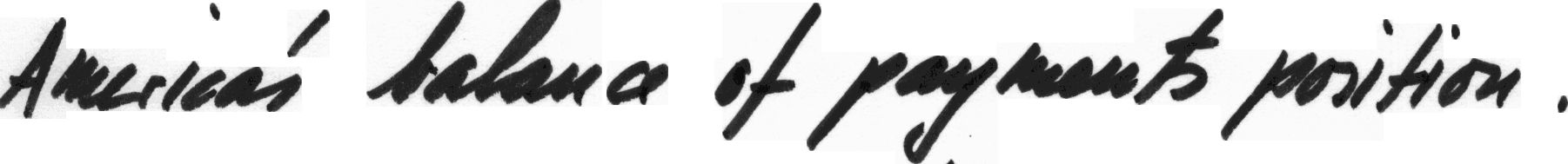}
    \end{subfigure}
    \begin{subfigure}[b]{\textwidth}
        \centering
        \includegraphics[width=0.99\linewidth]{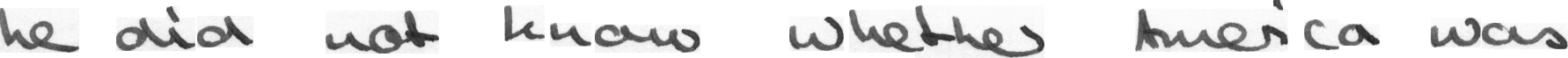}
    \end{subfigure}
    \caption{Line samples from the IAM dataset demonstrating unique challenges of the overall task.}
    \label{fig:lines}
\end{figure}

For Task 3, we were provided with a subset of the IAM dataset, which comprises handwritten lines authored by 657 distinct writers. Our specific subset includes 7,458 lines extracted from the 15,539 lines present in the complete dataset. These lines vary in size, length, and content, with examples illustrated in Figure \ref{fig:lines}. The original text, from which these lines have been transcribed, is sourced from the Lancaster-Oslo/Bergen Corpus of British English \cite{LOB_corpus}. Additionally, we were provided with a .txt file containing the ground truth for each line, ensuring accurate reference for our training and evaluation processes.

\section{Methods}
\label{sec:method}

\subsection{Task 1}

% \subsubsection{Line Segmentation}
% \label{sec:line-segmentation}
% \textit{As the team member who worked on this section dropped this course, we are unable to know its methodology.}

\subsubsection{Data Augmentation}

\paragraph{Data Collection} To increase the number of image-text pairs for training, we look into multiple online machine-typed sources. We set out to consider an online version of the Old Testament textual block with the Habakkuk font. Consequently, we identify (1) he.wiki\footnote{\url{https://he.wikisource.org/wiki/}}, and (2) thewaytoyahuweh\footnote{\url{https://downloads.thewaytoyahuweh.com/}} as the only sources that match our criterion. We filter punctuation from each textual block. In the end, we generate 40 textual files containing the textual version of the image-text training pairs.

\begin{figure}[h!]
    \centering
    \includegraphics[width=0.5\linewidth]{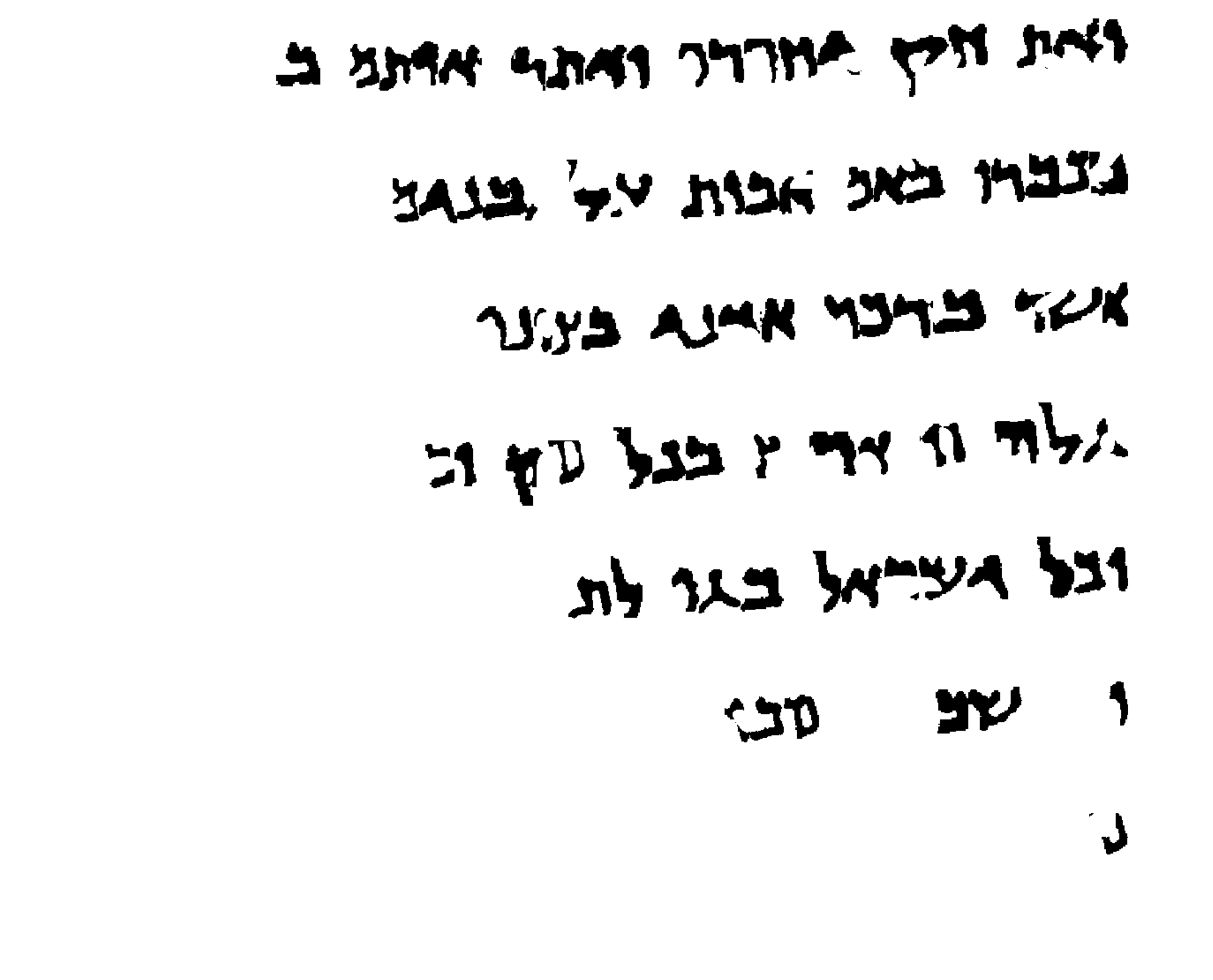}
    \caption{Example of an augmented image after performing Image Composition on a collected textual block.}
    \label{fig:task_1_noise_eg}
\end{figure}

\paragraph{Image Composition} To generate image composition, we take a blank canvas of $number\_of\_lines$ by $token\_size$ of the textual files. Later, we insert their corresponding character images from right to left. On each canvas, we set different parameters of Perlin noise\footnote{\url{https://rtouti.github.io/graphics/perlin-noise-algorithm}} to replicate the background and image texture of the test set. During the process, we assign values to layout parameters, \textit{letter spacing}, \textit{space width}, \textit{wave amplitude}, and \textit{wave frequency}  through experiments. Later, we perform our Line Segmentation technique\footnote{Due to ethical constraints, we are unable to disclose its implementation} on these images to generate augmented training samples for the model. We repeat these steps for each different version of a character we were provided in the task. Finally, we generate 9,438 line-segmented image-text pairs for the models to train on. A sample of the image of an image-text pair is provided in Figure \ref{fig:task_1_noise_eg}.

\subsubsection{Model}

\paragraph{TrOCR} We utilize TrOCR$_{Large}$ \cite{TrOCR}, which outperforms existing text recognition models through Transformer architecture. This architecture allows the model to understand images at the encoder level and generate text at the word level in the decoder. More specifically, the encoder was initialized with a pre-trained ViT model \cite{dosovitskiy2020image} while the decoder was initialized with RoBERTa \cite{liu2019roberta}. This model was trained on English datasets such as IIT-HWS \cite{krishnan2016generating} and the IAM Handwriting Database.

\paragraph{Kraken} We also experiment with Kraken\footnote{\url{https://kraken.re/main/index.html}} due to its architectural alignment with our pipeline. Specifically, this model performs line segmentation before character recognition. Furthermore, it has been trained on historical handwritten texts and is capable of processing right-to-left languages on an image, including Hebrew.

\subsection{Task 2}

\subsubsection{Data Processing}

\begin{figure}[h!]
    \centering
    \includegraphics[width=0.5\linewidth]{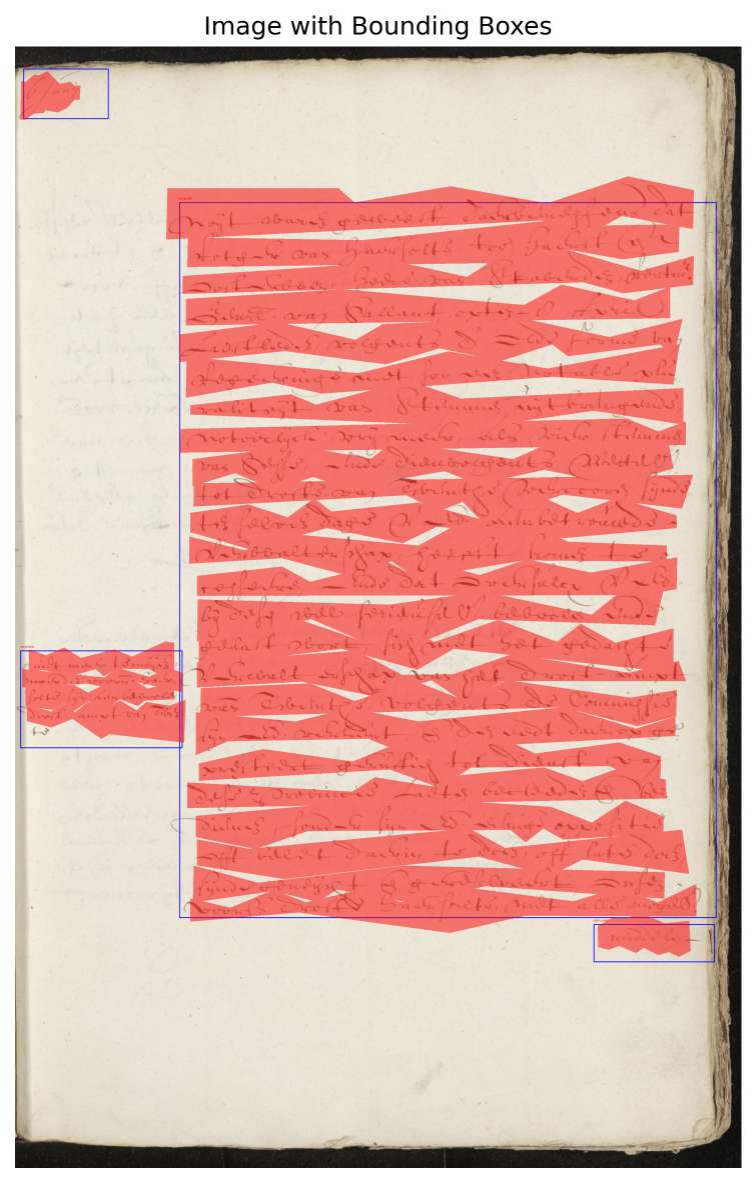}
    \caption{Preprocessed Resolution with Segmentation Mask and Boundary Boxes after page-separation preprocessing.}
    \label{fig:task_2_preprocess_image_with_train}
\end{figure}
To overcome resolution limitations, we decided to split double-page documents into single pages using the flood-fill algorithm on colored images. On each image, we initiated the algorithm at the coordinates (\verb|image_height| - (\verb|image_height|/3), (\verb|image_width|/2)) and moved left until we detected an \textit{R} value in the RGB color value falling below 200. This indicated a transition to a darker region, typically a border or margin, which we marked as the separation point between pages. In the case of single-page documents, our algorithm recognized the outer border and retained it as a single page.

We applied our method to all colored images in the training set, storing the identified separation \textit{x}-coordinates in a dictionary with file names as keys. These x-coordinates were then used to separate pages in the corresponding segmentation masks and binary images, adding the suffixes \textit{\_l} and \textit{\_r} to all related files for proper mapping. 

To align the bounding boxes of the coordinates accurately, we encountered three scenarios: (1) both ends of the x-axis of the box are on the left side, (2) both ends of the x-axis are on the right side, or (3) our identified \textit{x}-coordinate came in between the box. In case (3), we assessed whether one side of the box matched case (1) or (2) and set the other x-axis to our identified page-separated \textit{x}-coordinate.

Figure \ref{fig:task_2_preprocess_image_with_train} illustrates an example of case (3) on a colored image, where we set the x-axis of the \textit{marginally} and \textit{date} labels, respectively, to our identified \textit{x}-coordinate values to ensure no boxes were missed. We then repeated this algorithm on unlabeled color images to generate preprocessed binary and segmentation-mask images for the unlabeled set. Ultimately, we achieve a half-pixel dimension reduction for all images.

\subsection{Model}
% We utilize this model for both (name two task names)
%pretrained model: smp.DeepLabV3Plus

While reviewing the semantic mask images provided by the task organizers, we observed that each mask was distinguished by different colors corresponding to various labels. As a result, we decided to carry out both semantic segmentation and class label prediction tasks with color codes as our ground truth for the multi-class prediction task. We aimed to select a pre-trained semantic segmentation model capable of effectively converging on images with hard labels, i.e., distinguishing between foreground (text) and background. Consequently, the \verb|segmentation_models_pytorch.DeepLabV3Plus| model\footnote{\url{https://smp.readthedocs.io/en/v0.1.3/_modules/segmentation_models_pytorch/deeplabv3/model.html}} \cite{chen2017rethinking} effectively meets our requirements. Initially, we employ this model for semantic segmentation, and subsequently, we utilize it for multi-class prediction.

\subsection{Self-Supervised Learning}
% First, we load the best model
% We keep track of all new images that received average confidence among all labels, more than our determined confidence_threshold
% We append these new sets to the train_set and remove them from the evaluation set.

To overcome the limited training data, we leverage the 1,255 unlabeled images through self-supervised learning\footnote{For the rest of the paper, we will refer to \textit{self-supervised learning} when we utilize the unlabeled set for training and \textit{unsupervised learning} when we utilize only the 90 images provided to us by the task organizers}. For each individual task, we employ pseudo-labeling with the top-performing model on the entire unlabeled dataset. We then integrate images where the model demonstrates high average confidence into our training set, subsequently removing them from the unlabeled set. This process is repeated until we no longer see improvements in validation performance over a predetermined number of epochs. For the semantic segmentation task, this process continued once, while for the multi-class prediction task, it continued twice.

\subsection{Task 3}
%Preprocessing
% crnn-ctc-resnet34: architecture: RNN

\begin{figure}[h!]
    \centering
    \includegraphics[width=0.5\linewidth]{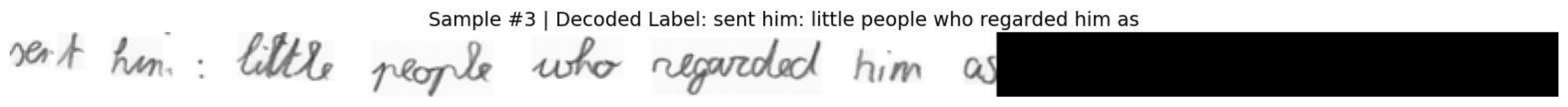}
    \caption{Example of our image preprocessing where we first perform padding and then blacking out the trailing whitespace region.}
    \label{fig:task_2_preprocess_image_with_train}
\end{figure}

To ensure the model effectively understands the images, we first resize the height while preserving the aspect ratio. We then pad each image to match the maximum width in the batch, carefully keeping track of the starting indices of the padding to eliminate trailing whitespace by drawing a black line over the padded areas, as shown in Figure \ref{fig:task_2_preprocess_image_with_train}.

We employ a Convolutional Recurrent Neural Network (CRNN) \cite{CRNN}, which consists of three main components: convolutional layers, recurrent layers \cite{memory1997sepp}, and a transcription layer. The convolutional layers at the base of the CRNN extract a feature sequence from each input image, which the recurrent network then uses to make predictions for each frame. The final transcription layer converts these per-frame predictions into a label sequence. Following the author's recommendations, we perform joint training using a single loss function.

\section{Experimental Setup}
\label{sec:exp-setup}

\subsection{Task 1}
%where was it experimented? (Python, Habrok)
% Evaluation Metric
% Train-Val Split
% Test Set
We implement our experiments on Nvidia A100 and V100 GPUs. We chose all the images composited from the first 33 textual files (i.e., 80\%) for the training and the rest for the validation set to avoid leakage. We report our performance on the two test samples, namely \textit{25-Fg001} and \textit{124-Fg004}, provided to us by the task coordinators. We evaluate the model's performance using the summation average of the Levenshtein distance ratio on both test samples. 

% Fully Trained
% Hyper-parameters
We perform end-to-end all-layer finetuning for both TrOCR and Kraken models. For the TrOCR model, we set special tokens of the decoder model to facilitate its synchronized learning with the encoder. For the training of the Kraken model, we set \verb|batch_size| to $8$ and \verb|min_delta| to $0.001$. Additionally, for both models, we set early stopping with a maximum epoch of 10.

\subsection{Task 2}
We implement our experiments on Nvidia V100 GPUs. We perform an 80-20 train-test split. Then, on the 80\% train set, we further perform an 85-15 train-val split. In all cases, we perform a random split with random\_state set to 42. We keep the same validation set and test set across unsupervised and self-supervised learning settings.

As we utilize the different color shades used for segmentation masking as the ground truth for the labels, we utilize \verb|Image.Resampling.NEAREST|\footnote{\url{https://pillow.readthedocs.io/en/stable/reference/Image.html}} to prevent interpolation of class labels. As the background was also comprised of its own color, we added another class (in total 9) during training. In this way, we also set all \textit{undefined} labels as the background labels. During loss propagation and evaluation, we did not consider the prediction of the background as a class label. During self-supervised learning, we normalize unlabel images by setting Mean and STD to values\footnote{\url{https://docs.pytorch.org/vision/stable/models.html}} suggested by the authors of \verb|DeepLabV3Plus| to improve transfer learning. As some unlabeled documents were empty, we only calculated and propagated loss for non-empty pages. We report our performances on the test set and evaluate using the mean of IOU\footnote{\url{https://smp.readthedocs.io/en/latest/metrics.html}} on the textual region for the semantic segmentation and 8 classes for the multi-class prediction task, respectively.

We utilize ResNet50 as our encoder, initializing its weights with ImageNet. The input channel count is set to 3. To mitigate overfitting on overrepresented classes in our multi-class prediction tasks, we implement weighted class computation with \( \epsilon \) set to \( 1e^{-9} \) to prevent division by zero. The formula we use is: 

\[
\frac{total\_pixels}{8 \times (foreground\_class\_pixel\_counts + \epsilon)}
\] 

where \( total\_pixels \) represents the sum of all class pixels, excluding \( background\_class\_pixel\_counts \). Throughout the experiments, we set all random seeds to 42. For both supervised and unsupervised training, the batch size is fixed at 2. We employ CrossEntropyLoss and the Adam optimizer for loss calculation and optimization, respectively. Due to time complexity considerations, we only experiment with a learning rate of 0.0001. In all cases, we stop training if the model's mean IOU does not improve on the validation set for five consecutive epochs. In the self-supervised training phase, we incorporate an unlabeled image into the training set if the model's average confidence is 0.70 or above.

\subsection{Task 3}
% evaluation metric: character-level and word-level accuracy

% weights=None
To maximize feature extraction capabilities, we initialize the \verb|ResNet34| architecture with empty weights before expanding the width of the feature map fed to the RNN. As the IAM images are grayscale, we modify the first convolution layer to accept grayscale images instead of RGB. Using \verb|adaptive_avg_pool2d|\footnote{\url{https://docs.pytorch.org/docs/stable/generated/torch.nn.functional.adaptive_avg_pool2d.html}}, we extract features from the $512$-dimensional output of the \verb|ResNet34|. This feature vector is then input into a bidirectional long short-term memory (Bi-LSTM) network with a dropout rate of $0.20$, followed by a \verb|log_softmax| output. For training, we utilize \verb|CTCLoss| with mean reduction and the \verb|AdamW| optimizer and set the learning rate to $0.001$. 

We evaluate our model on a test set comprising 100 randomly selected samples from the dataset, while an 80-20 split is applied to the remaining data for generating training and validation sets. We implement early stopping after $5$ epochs and assess performance using character-level\footnote{\url{https://lightning.ai/docs/torchmetrics/stable/text/char_error_rate.html}} and word-level\footnote{\url{https://lightning.ai/docs/torchmetrics/stable/text/word_error_rate.html}} error rates. Ultimately, our model was trained for a total of 24 epochs.

\section{Results}
\label{sec:results}

\subsection{Task 1}

We present the results of the two models fine-tuned on our dataset in Table \ref{tab:task_1_result_overall}. Notably, Kraken outperforms the TrOCR model by a ratio of $0.178$. We further present the performance of Kraken on individual test items and report them in Table \ref{tab:task_1_result_each_test_kraken}. Notice that the fine-tuned model finds the second test item, \textit{124-Fg004}, more challenging than the first test item \textit{25-Fg001}. 
Furthermore, we compare the results of TrOCR and Kraken between pre-trained and fine-tuned on our augmented dataset. Table \ref{tab:task1_finetune_vs_pretrained} illustrates that the TrOCR model demonstrates a significantly higher Laventish Distance Ratio compared to Kraken when both are fine-tuned on our augmented dataset. Specifically, the fine-tuned TrOCR model shows an improvement rate that is approximately five times greater than that of the fine-tuned Kraken model.

\input{tables/task_1_result_overall}
\input{tables/task_1_result_each_test_kraken}
\input{tables/task1_finetune_vs_pretrained}

%% Fine-tuned
% TRoCR (0.339) (no individual)
% Kraken (0.467, 0.426) (avg: 0.447)

\subsection{Task 2}
\input{tables/task_2_overall_result_table}

We present the result of the \verb|DeepLabV3Plus| model performance on both semantic segmentation and multi-class prediction tasks across unsupervised and self-supervised settings in Table \ref{tab:task_2_overall_result}. On both tasks, the model achieved performance gain with the self-supervised learning method. More specifically, we notice a very small performance gain of $0.002$ on the semantic segmentation task and a significant performance gain of $\approx$0.7 on the multi-class prediction task.
%% Semantic Segmentation
% supervised: 0.8437
% un-supervised: 0.8439

%% Multi-class Prediction
% supervised MIOU: 0.5892 
% un-supervised MIOU: 0.6506

\subsection{Task 3}
\input{tables/task_3_overall_result}

We report the loss, character error rate (CER), and word error rate (WER) of the best-performing model on the test set in Table \ref{tab:task_3_overall_result}. The results indicate that nearly 50\% of the predicted words and one-fifth of the predicted characters are incorrect.

\section{Discussion}
\label{sec:discuss}

\subsection{Task 1}
In this task, we recognize that the potential of our encoder-decoder architecture was not fully realized due to time constraints. It would have been intriguing to evaluate its performance against the relatively straightforward model we submitted. Furthermore, we did not perform thorough hyperparameter tuning or cross-validation, which might have enhanced the model’s performance. More time could have facilitated enhancements to our synthetic Dead Sea Scrolls dataset, which, while functional, still lacked the authenticity necessary for robust generalization. 

\subsection{Task 2}
Similar to Task 1, our efforts in hyperparameter tuning and cross-validation were limited. Our approach to data augmentation was minimal, confined to splitting pages in half, though employing more comprehensive augmentation techniques could have significantly bolstered the model's robustness. A notable missed opportunity was the failure to implement the classification head intended to follow the binary segmentation stage. Due to incomplete research and time limitations, we resorted to a multi-class segmentation method, which prevented us from exploring the potential benefits of the original modular setup. Additionally, our experimentation with the pre-trained models was limited, another limitation attributed to time constraints. Given our initial training of a binary segmentation model that was subsequently fine-tuned for multi-class segmentation, it would be valuable to conduct a comparative analysis with a multi-class model developed from scratch in future work.

\subsection{Task 3}
For this task, completing the training of the transformer-based model under development would have provided valuable insights through a direct comparison with the CRNN, particularly regarding how increased model complexity performs on a limited dataset. We also overlooked the opportunity to pretrain on a synthetic handwriting dataset, a strategy I had investigated but was unable to implement due to time limitations. The CRNN pipeline similarly lacked data augmentation, which could have further improved generalization. As with the previous tasks, we did not conduct hyperparameter tuning or cross-validation, which could have yielded important insights into model performance.

To address the observed word prediction error rate of 50\%, several strategies warrant consideration for future implementations: (1) the introduction of a post-correction module that enhances output accuracy, (2) the integration of pre-trained embeddings into the LSTM's embedding layer, or (3) the exploration of replacing the LSTM architecture with a model trained at the subword level. For instance, an effective enhancement could involve channeling the predicted output through a language model that corrects to produce grammatically accurate results. Moreover, utilizing pre-trained embeddings such as word2vec \cite{mikolov2013distributed} within the LSTM’s embedding layer could significantly advance the model’s performance on this task.

\section{Conclusion}
\label{sec:conclussion}
\subsection{Task 1}
Task 1 revealed the complexity of image synthesis for historical manuscripts, such as the Dead Sea Scrolls. Achieving an authentic representation of their unique texture and appearance demands careful data preparation and image processing, areas where our synthetic images, while functional, did not fully succeed. Despite this, our model performed robustly, likely benefiting from pretraining on a dataset of medieval handwritten Hebrew texts, which underscores the effectiveness of transfer learning in this context. This experience highlighted the difficulties inherent in working with limited and heterogeneous data, especially when generating synthetic samples that accurately capture real-world variations. Notably, non-transformer-based models also performed well, indicating that under data constraints, simpler architectures can be surprisingly effective. Overall, this task emphasizes the necessity for continued refinement of our methodologies and datasets to enhance the authenticity and performance of synthetic image generation in future research endeavors.

\subsection{Task 2}
In Task 2, our objective was to develop a two-stage pipeline that integrated a binary segmentation model with a subsequent classification head to effectively label distinct regions. This modular design was intended to enhance functional clarity within the model architecture. However, the implementation of this structure did not yield the anticipated performance, leading us to revert to a multi-class segmentation approach, which ultimately provided superior results. 

To address the challenge of limited labeled data, we employed pseudo-labeling techniques to augment our training set, resulting in notable improvements in model performance. This experience underscored a crucial insight: theoretical frameworks, while valuable, may not always translate seamlessly into practical applications, particularly when increased model complexity invites greater potential for error. Moving forward, it will be essential to balance innovative design with empirical validation to optimize performance outcomes.

\subsection{Task 3}
Task 3 highlighted the critical balance between model complexity and dataset size. Our attempts to integrate a transformer encoder between the CNN and RNN ultimately hindered performance, likely due to the inadequate size of our dataset. In contrast, we also explored the transformer-based model architecture, which demonstrated promising training metrics, underpinned by significant data augmentation, but was unable to be finalized by the deadline. The result underscored the importance of enhancing training data diversity, which can be as beneficial as increasing model complexity. This experience reinforces the notion that thoughtful data augmentation strategies can greatly improve model performance, even when working with constrained datasets.

\bibliographystyle{alpha}
\bibliography{sample}

\end{document}

%% file: tables/task_1_result_overall.tex
\begin{table}[h!]
    \centering
    
     \begin{tabular}{ l r } 
     \toprule
      \textbf{Model} & 
      \textbf{Laventish Distance Ratio}  \\
      % \textbf{\#Word/Ques. (En)} & 
      % \textbf{\#Word/Ques. (Bn)} \\
     \midrule
     \textbf{TrOCR} & 0.339  \\
     % & 180.51 ($\pm$ 93.86) & 156.30 ($\pm$ 82.96) \\
     
     \textbf{Kraken} & \textbf{0.447} \\
     % & 21.46 ($\pm$ 8.10) & 17.38 ($\pm$ 6.73)\\

     \bottomrule
     
    \end{tabular}
    \caption{Laventish Distance Ratios for each model evaluated on the two test sets, fine-tuned with augmented data.}
    \label{tab:task_1_result_overall}
\end{table}

% TRoCR (0.339) (no individual)
% Kraken (0.467, 0.426) (avg: 0.447)

%% file: tables/task_1_result_each_test_kraken.tex
\begin{table}[h!]
    \centering
    
     \begin{tabular}{ c r } 
     \toprule
      \textbf{Test Sample Name} & 
      \textbf{Laventish Distance Ratio}  \\
      % \textbf{\#Word/Ques. (En)} & 
      % \textbf{\#Word/Ques. (Bn)} \\
     \midrule
     \textbf{25-Fg001} & 0.467  \\
     % & 180.51 ($\pm$ 93.86) & 156.30 ($\pm$ 82.96) \\
     
     \textbf{124-Fg004} & 0.426 \\
     % & 21.46 ($\pm$ 8.10) & 17.38 ($\pm$ 6.73)\\

     \bottomrule
     
    \end{tabular}
    \caption{Evaluation of the fine-tuned Kraken model's Laventish Distance Ratio across each test item.}
    \label{tab:task_1_result_each_test_kraken}
\end{table}

% TRoCR (0.339) (no individual)
% Kraken (0.467, 0.426) (avg: 0.447)

%% file: tables/task1_finetune_vs_pretrained.tex
\begin{table}[h]
    \centering
    
     \begin{tabular}{ l l c } 
     \toprule
      \textbf{Model} & 
      \textbf{Training Method} & \textbf{Laventish Ratio}  \\
     \midrule
     \multirow{2}{5em}{TrOCR} & Pretrained & 0.171 \\ 
        & Finetuned & 0.339 (\textbf{+0.168}) \\ 
        \hline
        \multirow{2}{4em}{Kraken} & Pretrained & 0.418 \\ 
        & Finetuned & 0.447 (+0.029) \\ 

     \bottomrule
     
    \end{tabular}
    \caption{The performance comparison between the pre-trained and fine-tuned TrOCR and Kraken models. The TrOCR model demonstrates a greater performance improvement than the Kraken model when fine-tuned on our augmented training data compared to their pre-trained versions.}
    \label{tab:task1_finetune_vs_pretrained}
\end{table}

%% Semantic Segmentation
% supervised: 0.8437
% un-supervised: 0.8439

%% Multi-class Prediction
% supervised MIOU: 0.5892 
% un-supervised MIOU: 0.6506

%% file: tables/task_2_overall_result_table.tex
\begin{table}[h]
    \centering
    
     \begin{tabular}{ l r c } 
     \toprule
      \textbf{Task} & 
      \textbf{Learning Method} & \textbf{MIOU}  \\
     \midrule
     \multirow{2}{16em}{Semantic Segmentation} & Un-Supervised Learning & 0.8437 \\ 
        & Self-Supervised Learning & \textbf{0.8439} \\ 
        \hline
        \multirow{2}{16em}{Multi-class Prediction} & Un-Supervised Learning & 0.5892 \\ 
        & Self-Supervised Learning & \textbf{0.6506} \\ 

     \bottomrule
     
    \end{tabular}
    \caption{The performance of DeepLabV3Plus model on both Semantic Segmentation task and Multi-class Prediction task across self-supervised and unsupervised learning methods respectively.}
    \label{tab:task_2_overall_result}
\end{table}

%% Semantic Segmentation
% supervised: 0.8437
% un-supervised: 0.8439

%% Multi-class Prediction
% supervised MIOU: 0.5892 
% un-supervised MIOU: 0.6506

%% file: tables/task_3_overall_result.tex
\begin{table}[ht!]
    \centering
    
     \begin{tabular}{ l r } 
     \toprule
      \textbf{Evaluation Metric} & 
      \textbf{Performance} \\
      % \textbf{\#Word/Ques. (En)} & 
      % \textbf{\#Word/Ques. (Bn)} \\
     \midrule
     Loss & 0.926  \\
     % & 180.51 ($\pm$ 93.86) & 156.30 ($\pm$ 82.96) \\
     
     CER & 0.199 \\
     % & 21.46 ($\pm$ 8.10) & 17.38 ($\pm$ 6.73)\\

      WER & 0.528 \\

     \bottomrule
     
    \end{tabular}
    \caption{Loss, Character Error Rate (CER) and Word Error Rate (WER) of CRNN on our test set.}
    \label{tab:task_3_overall_result}
\end{table}

% TRoCR (0.339) (no individual)
% Kraken (0.467, 0.426) (avg: 0.447)

%% file: main.bbl
\newcommand{\etalchar}[1]{$^{#1}$}
\begin{thebibliography}{WWD{\etalchar{+}}20}

\bibitem[AA19]{mask_rcnn_instance_seg}
Abdullah Almutairi and Meshal Almashan.
\newblock Instance segmentation of newspaper elements using mask r-cnn.
\newblock In {\em 2019 18th IEEE International Conference On Machine Learning And Applications (ICMLA)}, pages 1371--1375, 2019.

\bibitem[BDPW21]{bao2021beit}
Hangbo Bao, Li~Dong, Songhao Piao, and Furu Wei.
\newblock Beit: Bert pre-training of image transformers.
\newblock {\em arXiv preprint arXiv:2106.08254}, 2021.

\bibitem[BM19]{DLA}
Galal~M Binmakhashen and Sabri~A Mahmoud.
\newblock Document layout analysis: a comprehensive survey.
\newblock {\em ACM Computing Surveys (CSUR)}, 52(6):1--36, 2019.

\bibitem[CCP22]{VerticalAttentionOCR}
Denis Coquenet, Cl{\'e}ment Chatelain, and Thierry Paquet.
\newblock End-to-end handwritten paragraph text recognition using a vertical attention network.
\newblock {\em IEEE Transactions on Pattern Analysis and Machine Intelligence}, 45(1):508--524, 2022.

\bibitem[CPSA17a]{DeepLabV3}
Liang-Chieh Chen, George Papandreou, Florian Schroff, and Hartwig Adam.
\newblock Rethinking atrous convolution for semantic image segmentation.
\newblock {\em arXiv preprint arXiv:1706.05587}, 2017.

\bibitem[CPSA17b]{chen2017rethinking}
Liang-Chieh Chen, George Papandreou, Florian Schroff, and Hartwig Adam.
\newblock Rethinking atrous convolution for semantic image segmentation.
\newblock {\em arXiv preprint arXiv:1706.05587}, 2017.

\bibitem[CW22]{itervm}
Xiaojie Chu and Yongtao Wang.
\newblock Itervm: iterative vision modeling module for scene text recognition.
\newblock In {\em 2022 26th International Conference on Pattern Recognition (ICPR)}, pages 1393--1399. IEEE, 2022.

\bibitem[CZP{\etalchar{+}}18]{DeepLabV3+}
Liang-Chieh Chen, Yukun Zhu, George Papandreou, Florian Schroff, and Hartwig Adam.
\newblock Encoder-decoder with atrous separable convolution for semantic image segmentation.
\newblock In {\em Proceedings of the European Conference on Computer Vision (ECCV)}, September 2018.

\bibitem[DBK{\etalchar{+}}20]{dosovitskiy2020image}
Alexey Dosovitskiy, Lucas Beyer, Alexander Kolesnikov, Dirk Weissenborn, Xiaohua Zhai, Thomas Unterthiner, Mostafa Dehghani, Matthias Minderer, Georg Heigold, Sylvain Gelly, et~al.
\newblock An image is worth 16x16 words: Transformers for image recognition at scale.
\newblock {\em arXiv preprint arXiv:2010.11929}, 2020.

\bibitem[GCG{\etalchar{+}}24]{Grosicki2024-za}
Emmanu{\`e}le Grosicki, Matthieu Carr{\'e}, Edouard Geoffrois, Emmanuel Augustin, Fran{\c c}oise Preteux, and Ronaldo Messina.
\newblock {RIMES}, complete, 2024.

\bibitem[Gra12]{graves2012connectionist}
Alex Graves.
\newblock Connectionist temporal classification.
\newblock In {\em Supervised sequence labelling with recurrent neural networks}, pages 61--93. Springer, 2012.

\bibitem[GS05]{bi_lstm}
Alex Graves and J{\"u}rgen Schmidhuber.
\newblock Framewise phoneme classification with bidirectional lstm networks.
\newblock In {\em Proceedings. 2005 IEEE International Joint Conference on Neural Networks, 2005.}, volume~4, pages 2047--2052. IEEE, 2005.

\bibitem[HZRS16]{resnet50}
Kaiming He, Xiangyu Zhang, Shaoqing Ren, and Jian Sun.
\newblock Deep residual learning for image recognition.
\newblock In {\em Proceedings of the IEEE Conference on Computer Vision and Pattern Recognition (CVPR)}, June 2016.

\bibitem[IDBN{\etalchar{+}}21]{transformerdata}
Andrei Ivanov, Nikoli Dryden, Tal Ben-Nun, Shigang Li, and Torsten Hoefler.
\newblock Data movement is all you need: A case study on optimizing transformers.
\newblock {\em Proceedings of Machine Learning and Systems}, 3:711--732, 2021.

\bibitem[KDN{\etalchar{+}}25]{CRNN}
Mahesh Kumar, Ajay Dureja, Rachna Narula, Ravi Arora, et~al.
\newblock Ocr-crnn (wbs): an optical character recognition system based on convolutional recurrent neural network embedded with word beam search decoder for extraction of text.
\newblock {\em International Journal of Information Technology}, pages 1--8, 2025.

\bibitem[Kie19]{kraken}
Benjamin Kiessling.
\newblock Kraken-a universal text recognizer for the humanities.
\newblock In {\em Digital Humanities 2019}, 2019.

\bibitem[KJ16]{krishnan2016generating}
Praveen Krishnan and CV~Jawahar.
\newblock Generating synthetic data for text recognition.
\newblock {\em arXiv preprint arXiv:1608.04224}, 2016.

\bibitem[Koo21]{koonce2021resnet}
Brett Koonce.
\newblock Resnet 34.
\newblock In {\em Convolutional neural networks with swift for tensorflow: image recognition and dataset categorization}, pages 51--61. Springer, 2021.

\bibitem[LBH15]{DL}
Yann LeCun, Yoshua Bengio, and Geoffrey Hinton.
\newblock Deep learning.
\newblock {\em nature}, 521(7553):436--444, 2015.

\bibitem[LCTS25]{HTR-VT}
Yuting Li, Dexiong Chen, Tinglong Tang, and Xi~Shen.
\newblock Htr-vt: Handwritten text recognition with vision transformer.
\newblock {\em Pattern Recognition}, 158:110967, 2025.

\bibitem[LLC{\etalchar{+}}23]{TrOCR}
Minghao Li, Tengchao Lv, Jingye Chen, Lei Cui, Yijuan Lu, Dinei Florencio, Cha Zhang, Zhoujun Li, and Furu Wei.
\newblock Trocr: Transformer-based optical character recognition with pre-trained models.
\newblock In {\em Proceedings of the AAAI conference on artificial intelligence}, volume~37, pages 13094--13102, 2023.

\bibitem[LOG{\etalchar{+}}19a]{roberta}
Yinhan Liu, Myle Ott, Naman Goyal, Jingfei Du, Mandar Joshi, Danqi Chen, Omer Levy, Mike Lewis, Luke Zettlemoyer, and Veselin Stoyanov.
\newblock Roberta: A robustly optimized bert pretraining approach.
\newblock {\em arXiv preprint arXiv:1907.11692}, 2019.

\bibitem[LOG{\etalchar{+}}19b]{liu2019roberta}
Yinhan Liu, Myle Ott, Naman Goyal, Jingfei Du, Mandar Joshi, Danqi Chen, Omer Levy, Mike Lewis, Luke Zettlemoyer, and Veselin Stoyanov.
\newblock Roberta: A robustly optimized bert pretraining approach.
\newblock {\em arXiv preprint arXiv:1907.11692}, 2019.

\bibitem[MB02]{iam}
U-V Marti and Horst Bunke.
\newblock The iam-database: an english sentence database for offline handwriting recognition.
\newblock {\em International journal on document analysis and recognition}, 5:39--46, 2002.

\bibitem[Mem97]{memory1997sepp}
Long Short-Term Memory.
\newblock Sepp hochreiter and j{\"u}rgen schmidhuber.
\newblock {\em Neural Computation}, 9(8):1735, 1997.

\bibitem[MT99]{Dead_sea_scrolls}
Florentino~Garc{\'\i}a Mart{\'\i}nez and Eibert~JC Tigchelaar.
\newblock {\em The Dead Sea Scrolls}.
\newblock Wm. B. Eerdmans Publishing Company, 1999.

\bibitem[RHGS16]{faster_r_cnn}
Shaoqing Ren, Kaiming He, Ross Girshick, and Jian Sun.
\newblock Faster r-cnn: Towards real-time object detection with region proposal networks.
\newblock {\em IEEE transactions on pattern analysis and machine intelligence}, 39(6):1137--1149, 2016.

\bibitem[SLG78]{LOB_corpus}
Johansson Stig, Geoffrey~N Leech, and Helen Goodluck.
\newblock Manual of information to accompany the lancaster-oslo: Bergen corpus of british english, for use with digital computers.
\newblock {\em (No Title)}, 1978.

\bibitem[TCJ22]{touvron2022deit}
Hugo Touvron, Matthieu Cord, and Herv{\'e} J{\'e}gou.
\newblock Deit iii: Revenge of the vit.
\newblock In {\em European conference on computer vision}, pages 516--533. Springer, 2022.

\bibitem[VSP{\etalchar{+}}17]{vaswani2017attention}
Ashish Vaswani, Noam Shazeer, Niki Parmar, Jakob Uszkoreit, Llion Jones, Aidan~N Gomez, {\L}ukasz Kaiser, and Illia Polosukhin.
\newblock Attention is all you need.
\newblock {\em Advances in neural information processing systems}, 30, 2017.

\bibitem[WWD{\etalchar{+}}20]{minilm}
Wenhui Wang, Furu Wei, Li~Dong, Hangbo Bao, Nan Yang, and Ming Zhou.
\newblock Minilm: Deep self-attention distillation for task-agnostic compression of pre-trained transformers.
\newblock In H.~Larochelle, M.~Ranzato, R.~Hadsell, M.F. Balcan, and H.~Lin, editors, {\em Advances in Neural Information Processing Systems}, volume~33, pages 5776--5788. Curran Associates, Inc., 2020.

\end{thebibliography}
